\documentclass[conference]{IEEEtran}
\IEEEoverridecommandlockouts
\usepackage{cite}
\usepackage{amsmath,amssymb,amsfonts}
\usepackage{algorithmic}
\usepackage{algorithm}
\usepackage{graphicx}
\usepackage{textcomp}
\usepackage{xcolor}
\def\BibTeX{{\rm B\kern-.05em{\sc i\kern-.025em b}\kern-.08em
    T\kern-.1667em\lower.7ex\hbox{E}\kern-.125emX}}

\newcommand{\mc}{\mathcal}
\newcommand{\ul}{\underline}

\newcommand{\tr}{\textrm}
\newcommand{\tb}{\textbf}

\begin{document}


\title{%
  A Graph Neural Network Model for Real-Time Gesture Recognition Based on sEMG Signals \\
\thanks{%
  This work is based on research supported by the National Science Foundation Grant \#2123635 and the University of Miami Provost Fellowship in Interdisciplinary Computing.}}

\author{%
  \IEEEauthorblockN{%
    Pragatheeswaran Vipulanandan, 
    Kamal Premaratne, 
    Manohar Murthi}
  \IEEEauthorblockA{%
    \textit{Department of Electrical and Computer Engineering} \\
    \textit{University of Miami} \\
    Coral Gables, Florida, USA \\
    pxv245@miami.edu, kamal@miami.edu, mmurthi@miami.edu}}


\maketitle


\begin{abstract}
For seemless control of advanced hand prostheses and augmented reality, accurate and immediate hand gestures recognition is essential. Surface electromyography (sEMG) signals obtained from the forearm are commonly employed for this purpose. In this paper, we present a novel approach for sEMG representation that utilizes graph networks which contain information about muscle activation patterns in the forearm. Based on these graph networks, we have developed a machine learning algorithm capable of real-time hand gesture recognition using a graph neural network. The algorithm's performance was evaluated using sEMG signals acquired from myoband, which has 8 electrodes placed around the forearm, involving 8 healthy subjects. The proposed method demonstrated an average classification accuracy of 99\%, surpassing the performance of state-of-the-art techniques. The average time for both graph construction and prediction stood at 48ms utilizing a M1 pro CPU, rendering the approach well-suited for real-time applications.
\end{abstract}


\begin{IEEEkeywords}
Graph neural networks, gesture recognition, sEMG signals
\end{IEEEkeywords}


\section{Introduction}


In the realm of human-computer interaction, real-time gesture recognition plays a pivotal role in enabling seamless communication between humans and machines. The ability to interpret human gestures accurately and swiftly holds immense potential for applications ranging from virtual reality (VR) and augmented reality (AR) systems to healthcare and robotics. Some of these systems utilize surface electromyogram (sEMG) signals which are captured from the forearm \cite{shenoy2008online, clement2011bionic}. The prevalent approach for gesture recognition involves extracting temporal and frequency domain features from the acquired sEMG recordings and subsequently classifying them using various learning algorithms, such as support vector machines (SVMs), linear discriminant analysis (LDA), and neural networks (NNs) \cite{chu2007supervised, tenore2007towards, chen2013pattern, raurale2018emg, de2020real}. 

While numerous studies have been conducted to accurately classify hand gestures from pre-recorded sEMG signals \cite{tenore2007towards, phinyomark2010evaluation, khushaba2012electromyogram}, only a limited number of research works have concentrated on \emph{real-time} hand gesture recognition using sEMG signals from the forearm \cite{crepin2018real, furui2019myoelectric, raurale2019emg}. The majority of existing methods utilize binning of sEMG signals and extracting specific features of the wave within each bin for classification purposes \cite{shima2010classification, khushaba2012electromyogram, crepin2018real, furui2019myoelectric}. Thus, the issue of computational complexity arising from continuous binning poses a challenge for real-time gesture prediction. In response to these challenges, this paper introduces an innovative approach harnessing the power of graph theory for real-time gesture recognition. The motivation for this research is rooted in the need for robust and efficient gesture recognition systems that can operate in real-time, allowing for natural and intuitive interactions between humans and digital interfaces.

Graph-based methods offer a versatile framework for modeling complex relationships and patterns within data, making them ideally suited for capturing the spatial and temporal dynamics of gestures \cite{yu2017spatio}. By representing gestures as dynamic graphs, we bridge the gap between raw sensor data and meaningful gesture representations. Each gesture is conceptualized as a graph, with nodes representing the electrodes on the human body and edges capturing the temporal dependencies and spatial correlations between the nodes  \cite{furui2019myoelectric}. Traditional graph kernel based methods such as shortest path kernel\cite{borgwardt2005shortest}, graphlet kernel \cite{shervashidze2009efficient}, random walk kernel \cite{vishwanathan2010graph}, Weisfelier-Lehman subtree kernel \cite{shervashidze2011weisfeiler}, propagation kernel \cite{neumann2012efficient}, and deep graph kernels \cite{yanardag2015deep} often fall short in terms of performance with smaller graph sizes, when compared with NN based methods \cite{zhou2023deep}.

In this paper, we propose a graph-based approach which builds upon recent advancements in graph neural networks (GNNs), graph convolution networks (GCNs) and graph signal processing (GSP) to enable the recognition of intricate gestures with minimal latency \cite{Gao20193D, chen2023gesture}. We employ the novel idea of constructing a graph network from time series sEMG signals that can represent the individual and mutual activation patterns of forearm muscles. We examine how this graph network could be employed to detect the time of gesture onset and also to classify different gesture types (for which we utilize a GNN). We explore its computational efficiency, real-time functionality, and accuracy, and demonstrate that that it surpasses the performance of state-of-the-art real-time gesture classification algorithms.


\section{Methodology}


\subsection{Overview}


In our study, we leverage the insights from previous research which has highlighted the crucial role that correlation between sEMG signal play in understanding complex electrical and chemical processes that occur in the nervous system dynamics \cite{de2020real}. Building upon this foundation, we adopt a graph-based approach to model the intricate relationships within our time series data. Specifically, we construct a graph representation where electrodes serve as nodes and the calculated correlations between the corresponding time series act as edges. By transforming our time series data into a graph structure based on these correlations, we gain a holistic perspective on the individual electrode activities and their inter-dependencies. This graph-based representation provides a powerful framework for analyzing the dynamic interactions and the correlations between sEMG signals and the gestures and enables us to unravel intricate patterns and gain deeper insights into the neuromuscular communication.


\subsection{Graph Generation}
\label{subsec:Preprocess}

\begin{figure*}[!t]
  \centering
  \includegraphics[width = 2.0\columnwidth]{%
    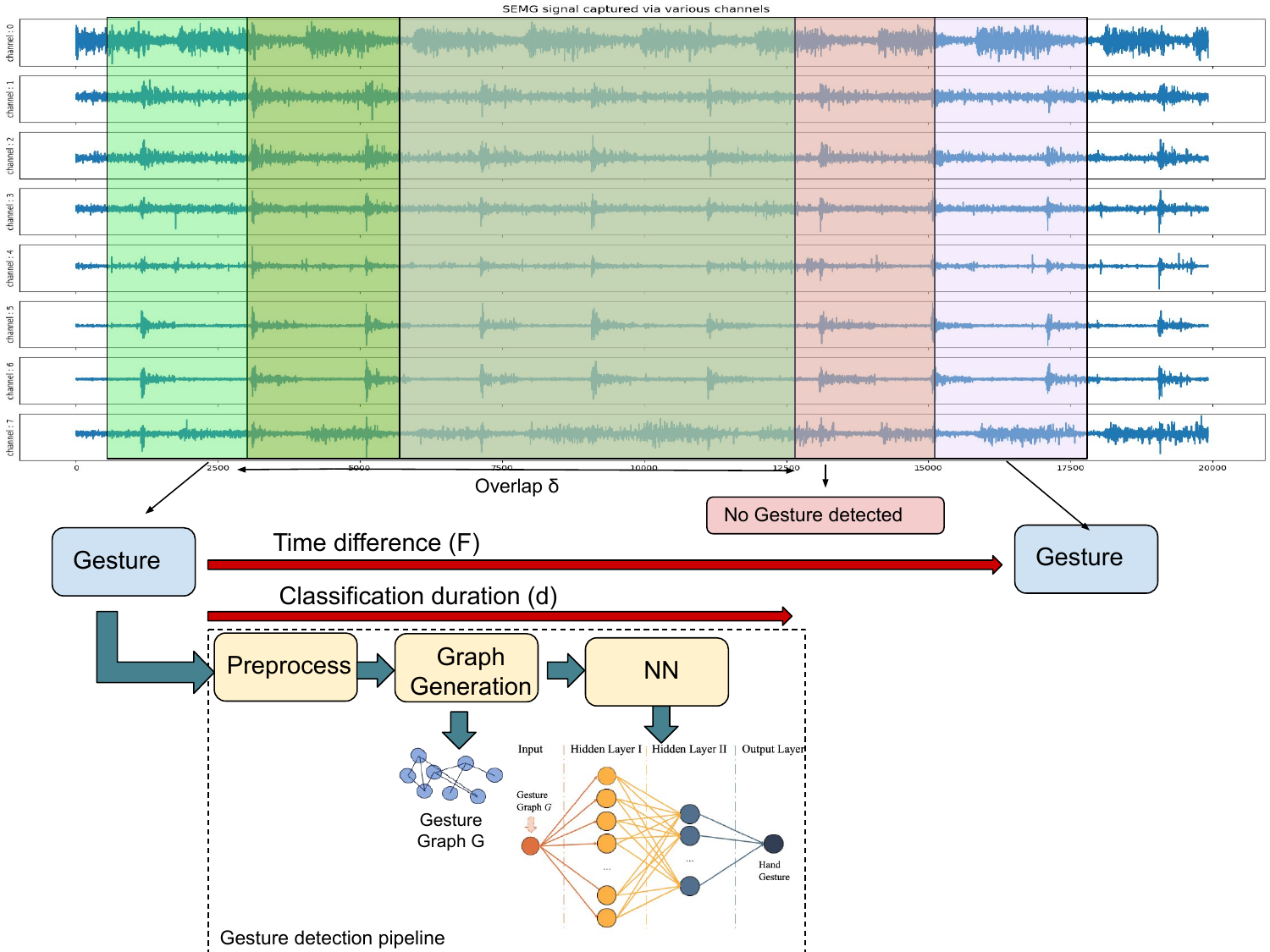}
  \caption{Pipeline of graph extraction from sEMG time series for gesture classification. Graphs are extracted only from those windows where a gusture is detected.} 
  \label{fig:gesture_predict}
\end{figure*}

Denote by $\{X_1(t),\, X_2(t),\, \ldots,\, X_M(t)\},\; t = 0, 1, 2, \ldots\,$, the surface sEMG signal time series recorded from $M$ channels. In the dataset that we explored (see Section~\ref{subsec:Dataset}), these time series correspond to sEMG signals produced by different muscle groups of human subjects. Here, $t$ is a non-negative integer that represents the sampling instant. Fig.~\ref{fig:gesture_predict} depicts the pipeline through which graphs are extracted from these sEMG time series for gesture classification.  

First, each continuous sEMG time series signal $X_i(\cdot)$ is partitioned into windows $W[n],\; n = 0, 1, 2, \ldots$, each window being of length $L$ samples and having an overlap percentage of $\delta$ samples with the next consecutive window. These windowed time series allow us to capture fine-grained temporal patterns within the signal. Let use denote the time series signal $X_i$ restricted to within the window $W[n]$ by $X_i[n]$, i.e., 
\begin{equation}
  X_i[n]
    = \{X_i[n](t) = X_i(nL + t),\; t = 0, 1, \ldots, L-1\}.
\end{equation}

Next, within each window $W[n]$, we calculate the Pearson correlation coefficient  between each pair of time series signals $X_i[n]$ and $X_j[n]$. This produces a $(M \times M)$-sized correlation matrix $\ul{P}[n]$ whose $(i, j)$-th element is the Pearson correlation coefficient between the time series $X_i[n]$ and $X_j[n]$ given by 
\begin{multline}
  \ul{P}[n]_{i,j} \\
    = \frac{%
        \displaystyle
        \sum_{t = 0}^{L - 1} 
        (X_i[n](t) - \overline{X_i[n]})\, (X_j[n](t) - \overline{X_j[n]})}{%
        \sqrt{%
          \displaystyle
          \sum_{t = 0}^{L - 1} 
          (X_i[n](t) - \overline{X_i[n]})^2} 
        \sqrt{%
          \displaystyle
          \sum_{t = 0}^{L - 1} 
          (X_j[n](t) - \overline{X_j[n]})^2}},
  \label{eq:pearson}
\end{multline}
where $\overline{X_i}[n]$ denotes the average signal amplitude of the time series $X_i[n](t),\; t = 0, 1, \ldots, L - 1$. The value of $\ul{P}[n]_{i,j}$ measures the linear relationship between $X_i[n]$ and $X_j[n]$; it takes values within the interval $[-1,\, +1]$, $+1$ indicating a perfect positive correlation, $-1$ indicating a perfect negative correlation, and $0$ indicating no correlation. 

Finally, we use this correlation matrix $\ul{P}[n]$, with its diagonal elements pinned at zero (so that $\ul{P}[n]_{i,i} = 0,\; i = 1, \ldots, M$), as the adjacency matrix of a graph $\mathcal{G}[n] = (\mathcal{V}[n],\, \mathcal{E}[n])$. Here, $\mathcal{V}[n]$ represents the set of nodes and $\mathcal{E}[n]$ represents the set of edges with the calculated Pearson correlation coefficients constituting the edge weights. Note that, $\mathcal{G}[n]$ is undirected (because $P[n]_{i,j} = P[n]_{j,i}$) and is free of self-loops (because $P[n]_{i,i} = 0$). Thus, it is a simple, fully connected undirected graph \cite{Newman2018_Book} associated with the $M$ time series falling within the window $W[n]$. 


\subsection{Event Detection}


With the undirected graph sequence $\mc{G}[n],\; n = 0, 1, 2, \ldots$, generated from the time series in hand, our objective is to identify meaningful temporal segments where significant gestures had occurred. Akin to what is used in \cite{de2020real}, we  achieve this by employing the following event detection measure: 
\begin{equation}
  \Delta[n, \tau]
    = \sqrt{%
        \sum_{i = 1}^M 
        \sum_{t = 0}^{L-1}
        \left(
          X_i[n](t) - X_i[n - \tau](t)
        \right)^2}, 
  \label{eq:error}
\end{equation}
where $X_i[n - \tau](t)$ is the time series falling within the earlier window $W[n - \tau]$, for some positive integer $\tau$. Note that, one may view this as the Fr{\"o}benius norm of a certain `error' matrix. To be specific, consider the $(M \times L)$-sized matrix  
\begin{multline}
  \ul{X}[n](0:L - 1) \\
    = \begin{bmatrix}
        X_1[n](0) & X_1[n](1) & \cdots & X_1[n](L - 1) \\
        X_2[n](0) & X_2[n](1) & \cdots & X_2[n](L - 1) \\
        \vdots & \vdots & \ddots & \vdots \\
        X_M[n](0) & X_M[n](1) & \cdots & X_M[n](L - 1)
      \end{bmatrix},
\end{multline}
for  $n = 0, 1, \ldots$. Then,
\begin{equation}
  \Delta[n, \tau]
    = \Vert 
        \ul{X}[n](0:L - 1) - \ul{X}[n - \tau](0:L - 1)
      \Vert_F.
\end{equation}
The $i$-th row of $\ul{X}[n](0:L - 1)$ corresponds to the samples $0,1, \ldots, L - 1$ of the time series $X_i$ that lie within window $W[n]$; the $i$-th row of $\ul{X}[n - \tau](0:L - 1)$ corresponds to the time series $X_i$ that lie within window $W[n - \tau]$. 
        
We flag the occurrence of a gesture event within only those windows $W[n]$ where $\Delta[n, \tau]$ exceeds a predefined threshold. Similar to the strategy employed in \cite{de2020real}, for event detection purposes, we consider only the $L$-length time series $X_i[n],\; i = 1, \ldots, N$, falling within these windows $W[n]$ where a gesture event had occurred. This focused approach ensures that our predictions are based on meaningful temporal segments characterized by significant events thus avoiding regions of low signal-to-noise ratio.


\subsection{Classification}


\subsubsection{Training}
\label{subsec:Training}


The graphs associated with the windows where an event had occurred, together with the corresponding hand gesture labels, serve as the training data for our classification algorithm. They encapsulate the correlation between signal channels thus reflecting muscle activity related to specific hand gestures. A straightforward GNN architecture, which includes two fully connected layers with a ReLU activation function, was then trained on this training data set.


\subsubsection{Testing}


The trained GNN is then used to classify hand gestures. It is worthwhile mentioning that leveraging the graph structure to propagate and sustain the information between channels allows even a simple network to understand the context and relationship in the data \cite{battaglia2018relational}. 


\section{Experiment}


\subsection{Dataset}
\label{subsec:Dataset}


\begin{figure}[!t]
 .\begin{center}
  \includegraphics[width = 1.0\columnwidth]{%
    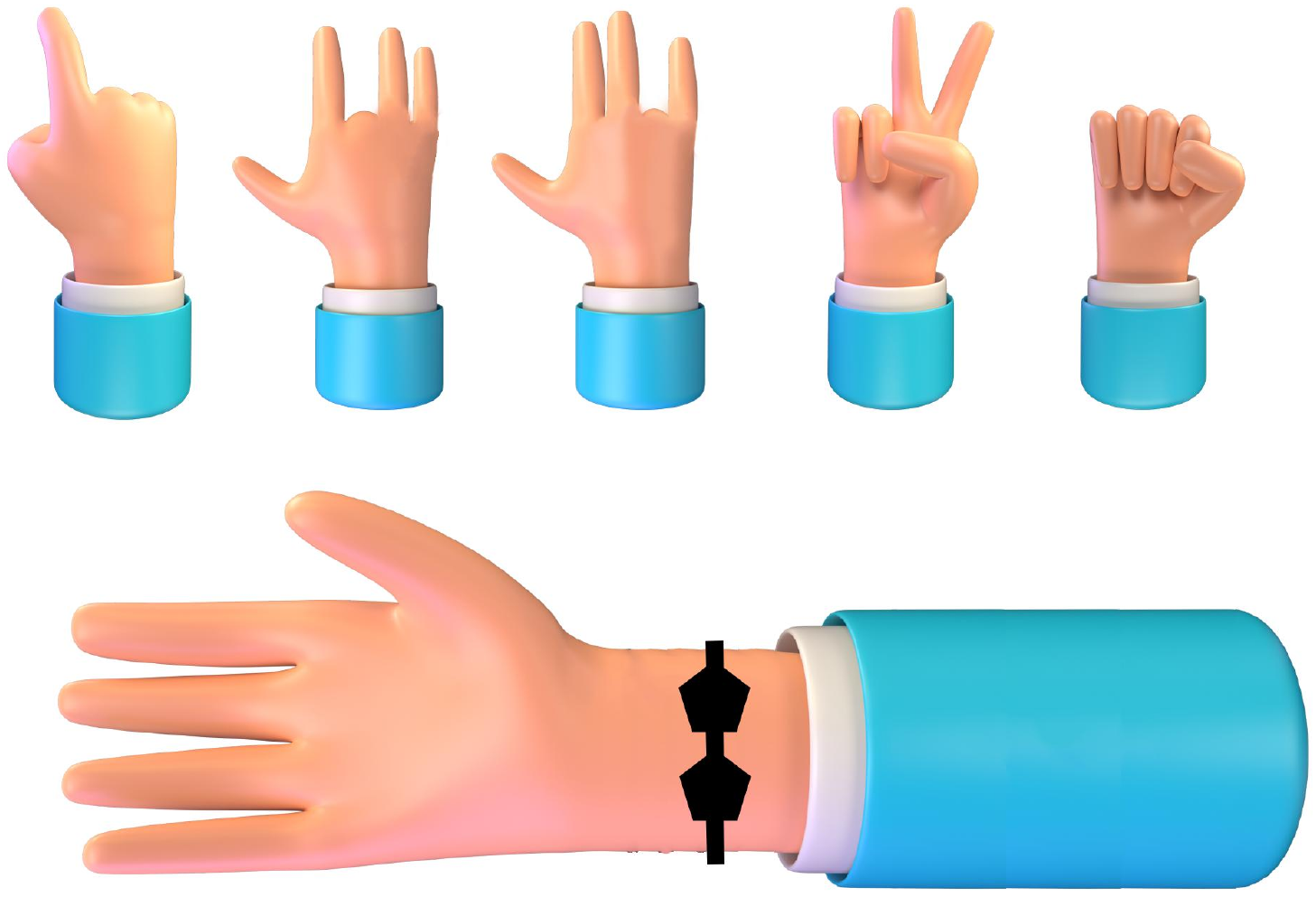}
  \caption{%
  Top row left-to-right: sEMG signals of the dataset captures the 5 hand gestures, pointer, middle finger flexion, ring finger flexion, v-flexion, and hand closure. Bottom row: the hand in the neutral position with Myo band placed in forearm.} 
  \label{fig:guesture_variety}
  \end{center}
\end{figure}

The dataset, sourced from \cite{de2020real}, captures sEMG signal activity of $C = 5$ gestures, pointer, middle finger flexion, ring finger flexion, v-flexion, and hand closure (see Fig.~\ref{fig:guesture_variety}). The signals were recorded at a sampling rate of $200\, \tr{Hz}$ from $K = 8$ healthy subjects (4 males and 4 females, aged $25 \pm 2$ years) using a Myo band (Thalamic Labs, Canada). During data collection, the subjects were asked to perform 20 repetitions of each gesture with their right hand. A single time series (roughly 20,000 samples) is generated by one subject repeatedly holding one gesture for a period of $5\, \tr{s}$ followed by a resting period of $5\, \tr{s}$. Each subject performs this task with 8 electrodes for each gesture, thus producing 40 time series. With $K = 8$ subjects participating in the study, we thus have $40 K = 320$ time series in total.


\subsection{Parameters}


To extract the graphs from the time series, we used a window width of $L = 80\, \tr{samples}$ (corresponding to $0.4\, \tr{s}$), an overlap of $\delta = 95\%$ between consecutive windows, and for computing the event detection measure in \eqref{eq:error} a window delay of $\tau = 5$ windows (corresponding to 20 samples). To select the event detection threshold, we followed the procedure that is utilized in \cite{de2020real} where the threshold employed is subject-dependant. 

Let us denote by $T_k$ the threshold employed for the time series produced by the $k$-th, where $k = 1, \ldots, K = 8$. To get $T_k$, we first compute the event detection measure $\Delta_{k, c}[n, \tau]$ separately for each gesture type. Here, $k,\; k = 1, \ldots, K = 8$, identifies the subject and $c,\; c = 1, \ldots, C = 5$, identifies the gesture. Let $\sigma^2(\Delta_{k, c}[\tau])$ denote the variance of $\Delta_{k, c}[n, \tau]$ over the windows, i.e., 
\begin{equation}
  \sigma^2(\Delta_{k, c}[\tau])
    = \frac{1}{N}
      \sum_{n = 0}^{N - 1}
      (\Delta_{k, c}[n, \tau] - \overline{\Delta}_{k, c}[n, \tau])^2,
\end{equation}
where $N$ denotes the number of windows and $\overline{\Delta}_{k, c}[n, \tau]$ denotes the average over the windows. These variances are then used to determine the subject-dependant event detection threshold as \cite{de2020real}
\begin{equation}
  T_k 
    = \frac{4}{C}
      \sum_{c = 1}^C 
      \sqrt{\sigma^2(\Delta_{k, c}[\tau])},\;
      k = 1, \ldots, K = 8.
\end{equation}
For subject $k$, an event is considered to occur within a specific window if $\Delta[n, \tau] \geq T_k$, i.e., the event detection metric in \eqref{eq:error} exceeds the threshold $T_k$. The entire window, in which an event is detected, is then included in the training dataset, while segments where no event is detected are ignored. 


\subsection{Architecture}


After extracting a training dataset of graphs for each of the subjects and their respective hand gestures, the GNN (implemented using PyTorch) was trained with the stochastic gradient descent optimizer using the following parameters: learning rate of 0.01 for 100 epochs, and 64 and 2048 neurons in the first fully and second fully connected layers, respectively. 


\subsection{Training and Real-Time Prediction}


The GNN model was evaluated using a randomly selected 20\% extracted from the original dataset. This testing dataset had an average of 927 graphs per class for an overlap of $\delta = 90\%$.

The classifier, after training, was employed in real-time for predicting gestures. The pipeline through which the gestures are classified using the trained model appears in Fig.~\ref{fig:gesture_predict}. The pseudo-code of the prediction algorithm is in Algorithm~\ref{alg:algorithm1}.

\begin{algorithm}[!t]
    \caption{Real-time gesture detection pipeline}
    \label{alg:algorithm1}
    \begin{algorithmic}
        \STATE  \tb{$subject \to k$}
        \STATE  \tb{$window \to n$}
        \STATE  Fetch previous $L$ samples $\ul{X}[n](0:L - 1)$; 
        
        \STATE \tb{Calculate event detection}
        \STATE $\Delta[n, \tau] \leftarrow$ event likelihood metric based on \eqref{eq:error}
        
        \IF {$\Delta[n, \tau] \geq$ event threshold ($T_k$)}
            \STATE \tb{Generate graph network}
            \STATE $\mathcal{G}[n] \leftarrow$ graph network based on matrix $\ul{P}[n]$
            \STATE prediction $\leftarrow \mathrm{ GNN}(\mathcal{G}[n])$
            
        \ENDIF
        \STATE \tb{Wait for $t_0$ number of new samples}
        \STATE $n \leftarrow n + 1$
    
    \end{algorithmic}
\end{algorithm}

Initially, the framework begins with the acquisition of new data, which is seamlessly captured from the Myo band. The framework awaits the accumulation of $t_0 = 20$ (equivalent to $F = 0.1\, \tr{s}$) new samples before initiating the subsequent phase of processing.  This intentional waiting period ensures that the framework captures gestures during transitions, therefore enhancing the precision and efficacy of our process.

The model prediction pipeline was configured to activate upon detection of a gesture event. Once an event was detected, the window of samples was utilized to generate a graph, utilizing the preprocessing described in Section~\ref{subsec:Preprocess}. The generated graph is inferenced using the trained classifier to identify the gesture. The time taken for the end-to-end process of detecting a gesture from the data is denoted as $d$. We demonstrate that the time required to process and classify a gesture is significantly shorter than the time the framework awaits the accumulation of $t_0$ samples, thereby ensuring real-time applicability.


\section{Results and Discussion}


The classification accuracies obtained from the proposed real-time gesture recognition algorithm and a comparison with state-of-the-art methods described in \cite{crepin2018real, furui2019myoelectric, de2020real} appear in Table~\ref{tab:1}. Crepin, et al. \cite{crepin2018real} employed a continuous binning approach, applying a LDA classifier to each bin while treating channels as independent entities. In contrast, Furui, et al. \cite{furui2019myoelectric} utilized a muscle synergy-based approach, conducting classification for each bin through a recurrent log-linearized Gaussian mixture network. On the other hand, Ashwin, et al. \cite{de2020real} utilized a temporal muscle activation (TMA) image based approach, which utilized a CNN for classification. 

Our method appears to offer superior accuracy in recognizing gestures compared to these techniques. In comparison to Crepin, et al. \cite{crepin2018real}, our method exhibited higher classification accuracy across all gestures, possibly attributed to our consideration of both individual and mutual muscle activations. When compared with Ashwin, et al. \cite{de2020real}, utilizing correlation between time series instead of the TMA mapping appears to yield better results. Additionally, the data involves complex relationships and dependencies that are not easily captured by regular grid structures. It is easier to capture these dependencies via a graph network based approach (a la GNN) than a CNN based approach. 

\begin{figure}[!t]
  \centering
  \includegraphics[width = 0.75\columnwidth]{%
    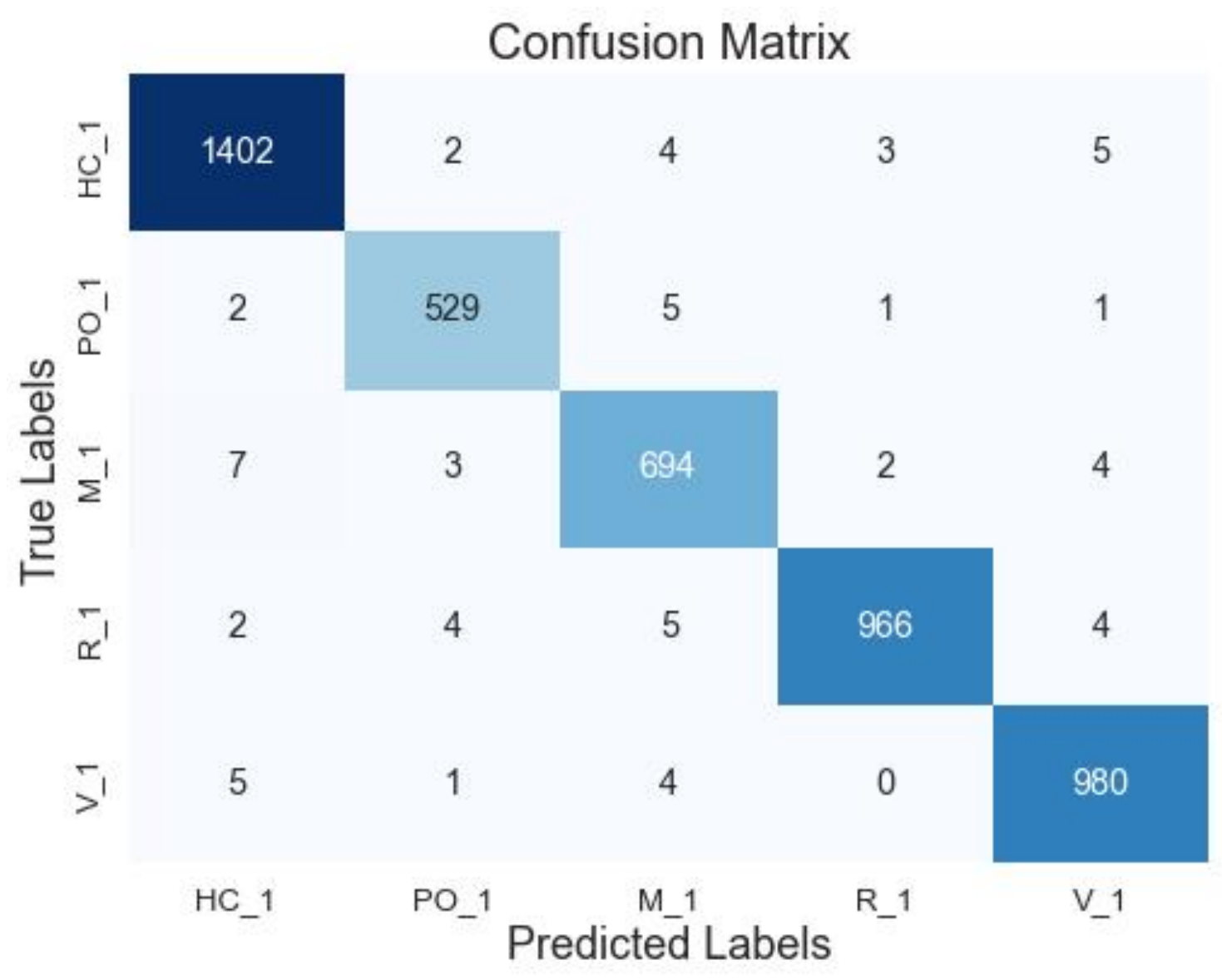}
  \caption{Confusion matrix associated with window overlap parameter $\delta=90\%$.} 
  \label{fig:confusion_matrix}
\end{figure}

In Fig \ref{fig:confusion_matrix}, we report the confusion matrix corresponding to our approach. The proposed method exhibited an average classification accuracy above 98\%. Notably, the model achieved this performance even in the presence of a significant class imbalance problem.

\begin{figure}[!t]
  \centering
  \includegraphics[width = 0.95\columnwidth]{%
    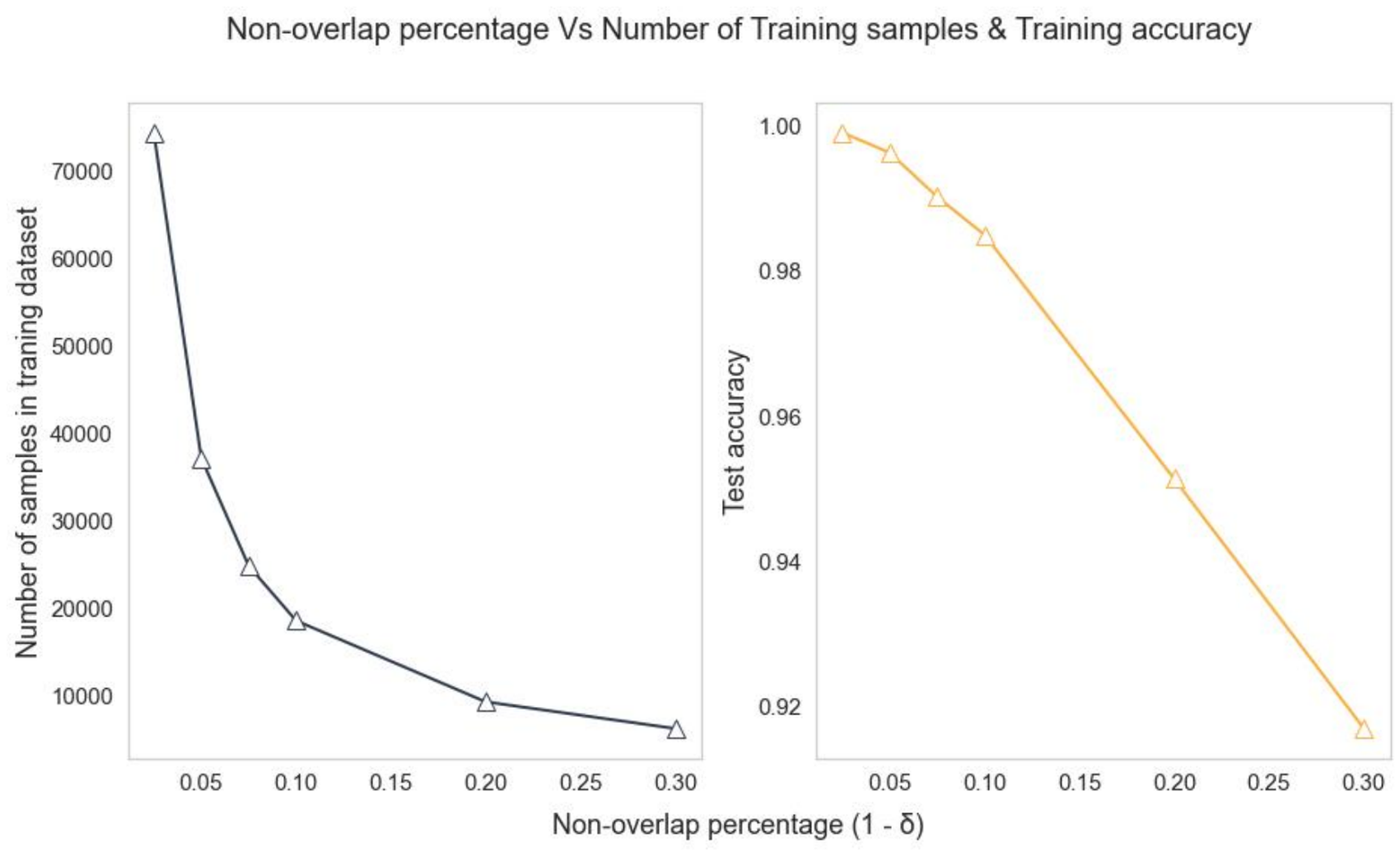}
  \caption{Variation of number of training samples (left) and accuracy (right) with respect to window non overlap parameter $1 - \delta$.} 
  \label{fig:delta_comparision}
\end{figure}

To systematically test the robustness of the algorithm, we studied the model's performance with different window overlap (i.e, $\delta$) values. With decreasing $\delta$, the number of graphs being produced is reduced meaning that the GNN has less training data from which to learn. Fig.~\ref{fig:delta_comparision} shows the training sample size and the test accuracies which were yielded. The proposed method demonstrates significant performance even in limited training data scenarios (i.e., lower $\delta$ values). In contrast. the work in Ashwin, et al. require a larger overlap of consecutive windows ($\delta = 97.5\%$) to maximize their training dataset. This highlights the GNN's capacity to generalize effectively even with limited training data. It could be noted that our approach surpasses testing accuracy of 99\% when training dataset was maximized via the appropriate selection of $\delta$ ($\delta = 97.5\%$).

To test the model's ability to distinguish between classes and to illustrate the trade-off between the true positive rate (sensitivity) and the false positive rate, we employed micro average Receiver Operating Characteristic (ROC) analysis with respect to the window overlap parameter $\delta$. The Area Under the Curve (AUC) is seen to be affected only slightly with decreasing $\delta$. This provides a clear measure of the GNN's responsiveness to changes in overlap, offering insights into model robustness. It is noteworthy that, when $\delta$ was reduced to 70\%, only 15\% of the maximum attainable training data samples (i.e., graphs) were produced. Despite this significant reduction, an AUC score of 0.967 was attained, highlighting the GNN's capacity to sustain high performance levels even with a diminished training dataset.

\begin{figure}[!t]
  \centering
  \includegraphics[width = 0.95\columnwidth]{%
    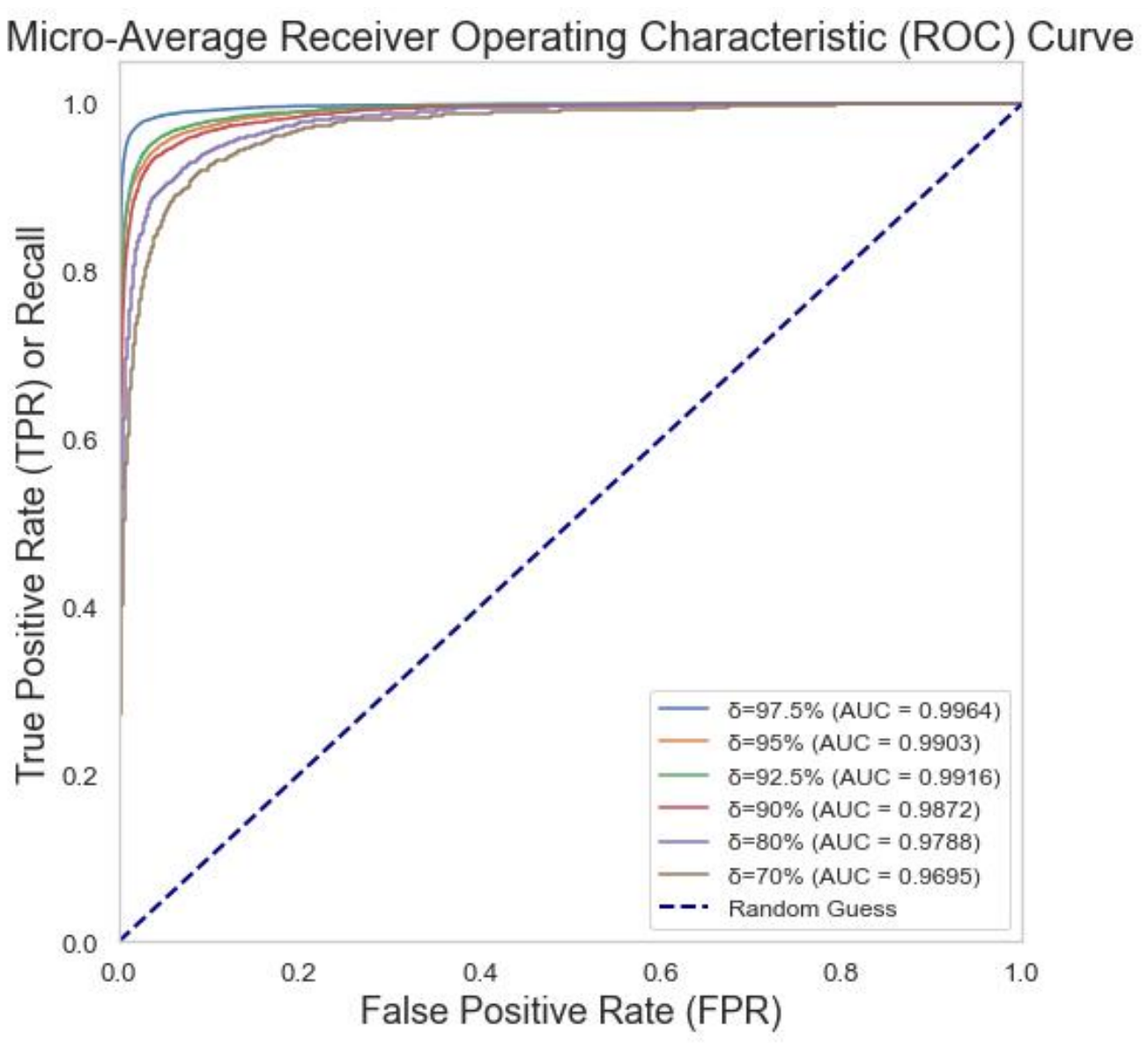}
  \caption{ROC curves with respect to window overlap parameter $\delta$.} 
  \label{fig:roc_curve}
\end{figure}

The average computation time of the proposed method was $48\, \tr{ms}$ ($3\, \tr{ms}$ for graph generation and $45\, \tr{ms}$ for prediction). The CPU usage was 39\% for gesture prediction when executed on a computer featuring a M1 pro CPU. Considering that this computation time was notably shorter than the duration between two successive events (denoted as $F$ in Fig.~\ref{fig:gesture_predict}), the algorithm appears to support real-time execution.

\begin{table}[!htbp]
  \caption{Classification Accuracy (\%)}
  \centering
  \begin{tabular}{l ccccc}
    \hline
    \hfil\tb{Hand Gestures} 
      & \tb{\cite{crepin2018real}}
      & \tb{\cite{furui2019myoelectric}}
      & \tb{\cite{de2020real}}
      & \multicolumn{2}{c}{\tb{Proposed Method}} \\ 
    \hline
    Overlap Percentage($\delta$) 
     & - & - & - & 90\% & 97.5\% \\
    \hline
    Middle Flexion 
      & 81.90 & 91.30 & 96.67 & \tb{97.40} & \tb{99.76}\\
    \hline
    Ring Flexion 
      & 93.50 & -- & 94.58 & \tb{98.37} & \tb{99.95}\\
    \hline
    Hand Closure 
      & 77.87 & 97.01 & 93.75 & \tb{98.94} & \tb{99.86}\\
    \hline
    V-Flexion 
      & -- & 95.06 & 92.91 & \tb{98.89} & \tb{99.95}\\
    \hline
    Pointer 
      & 80.85 & 97.27 & 92.50 & \tb{97.40} & \tb{99.91}\\
    \hline\hline
    Average 
      & 83.53 & 95.16 & 94.08 & \tb{98.19} & \tb{99.89}\\
    \hline
  \end{tabular}
  \label{tab:1}
\end{table}


\section{Conclusion}


In this study, we have presented a novel technique for generating undirected weighted graphs based on the interrelationships between sEMG signals. Our method was applied to identify the onset of gestures and successfully recognize five specific hand gestures in real-time. The accuracy achieved by our approach seems to be higher than the performance offered by state-of-the-art methods, thus showcasing the effectiveness of graph-based approaches. The method also showcases robustness across training datasets of various sizes. 

Looking forward, it might be possible to develop real-time fluid gesture recognition, representing a significant leap forward for individuals with disabilities. One may also explore regions where events went unnoticed to improve gesture detection, even in situations with a sub-optimal signal-to-noise ratio.

 
\bibliographystyle{%
  IEEEtran}

\bibliography{%
  LibraryPremaratne_Books,%
  LibraryPremaratne_Conferences,%
  LibraryPremaratne_Journals,%
  LibraryPremaratne_Other,%
  LibraryPremaratne_Datasets}


\end{document}